\documentclass{article}
\usepackage{spconf,amsmath,graphicx}
\usepackage{multirow}

\title{Cross-Modal Message Passing for Two-stream Fusion}
\name{Dong Wang \qquad Yuan Yuan \qquad Qi Wang \thanks{$^{*}$Corresponding author. This work was supported by the National Key Research and Development Program of China under Grant 2017YFB1002202, National Natural Science Foundation of China under Grant 61773316 and 61379094, Fundamental Research Funds for the Central Universities under Grant 3102017AX010, and the Open Research Fund of Key Laboratory of Spectral Imaging Technology£¬Chinese Academy of Sciences.}}
\address{School of Computer Science and Center for OPTical IMagery Analysis and Learning (OPTIMAL), \\
Northwestern Polytechnical University, Xi'an 710072, Shaanxi, P. R. China\\}

\begin{document}
%
\maketitle
\begin{abstract}
Processing and fusing information among multi-modal is a very useful technique for achieving high performance in many computer vision problems. In order to tackle multi-modal information more effectively, we introduce a novel framework for multi-modal fusion: Cross-modal Message Passing (CMMP). Specifically, we propose a cross-modal message passing mechanism to fuse two-stream network for action recognition, which composes of an appearance modal network (RGB image) and a motion modal (optical flow image) network. The objectives of individual networks in this framework are two-fold: a standard classification objective and a competing objective. The classification object ensures that each modal network predicts the true action category while the competing objective encourages each modal network to outperform the other one. We quantitatively show that the proposed CMMP fuse the traditional two-stream network more effectively, and outperforms all existing two-stream fusion method on UCF-101 and  HMDB-51 datasets.
\end{abstract}
\section{Introduction}
\label{sec:intro}
Video-based action recognition is an important problem in computer vision which has attracted great attention from the academic community \cite{NIPS2014_5353,wang2013action,wang2013motionlets,yue2015beyond}. It has various applications such as video surveillance, human-computer interface, and behavior analysis. Unlike recognition in static images, the motion dynamics is another crucial aspect for action recognition except visual appearance.

Recently, Convolutional Networks (ConvNets) \cite{lecun1998gradient} have witnessed great success in computer vision tasks such as image classification \cite{krizhevsky2012imagenet}. Researchers have applied ConvNets to solve the problem of video-based action recognition \cite{karpathy2014large,tran2015learning,zhang2016real} and achieved remarkable performance on public action recognition datasets. Deep ConvNets have been shown to have an extraordinary ability to extract visual appearance features from raw pixels, which is main component in image classification. In order to handle videos, many previous works represent the motion dynamics of video with an image format such as optical flow image, which can utilize deep ConvNet to extract powerful representations. Following this idea, the two-stream architecture \cite{simonyan2014two} incorporates motion information by training separate ConvNets for both appearance in still images and stacks of optical flow images.

\begin{figure}
\centering
\includegraphics[width=.48\textwidth]{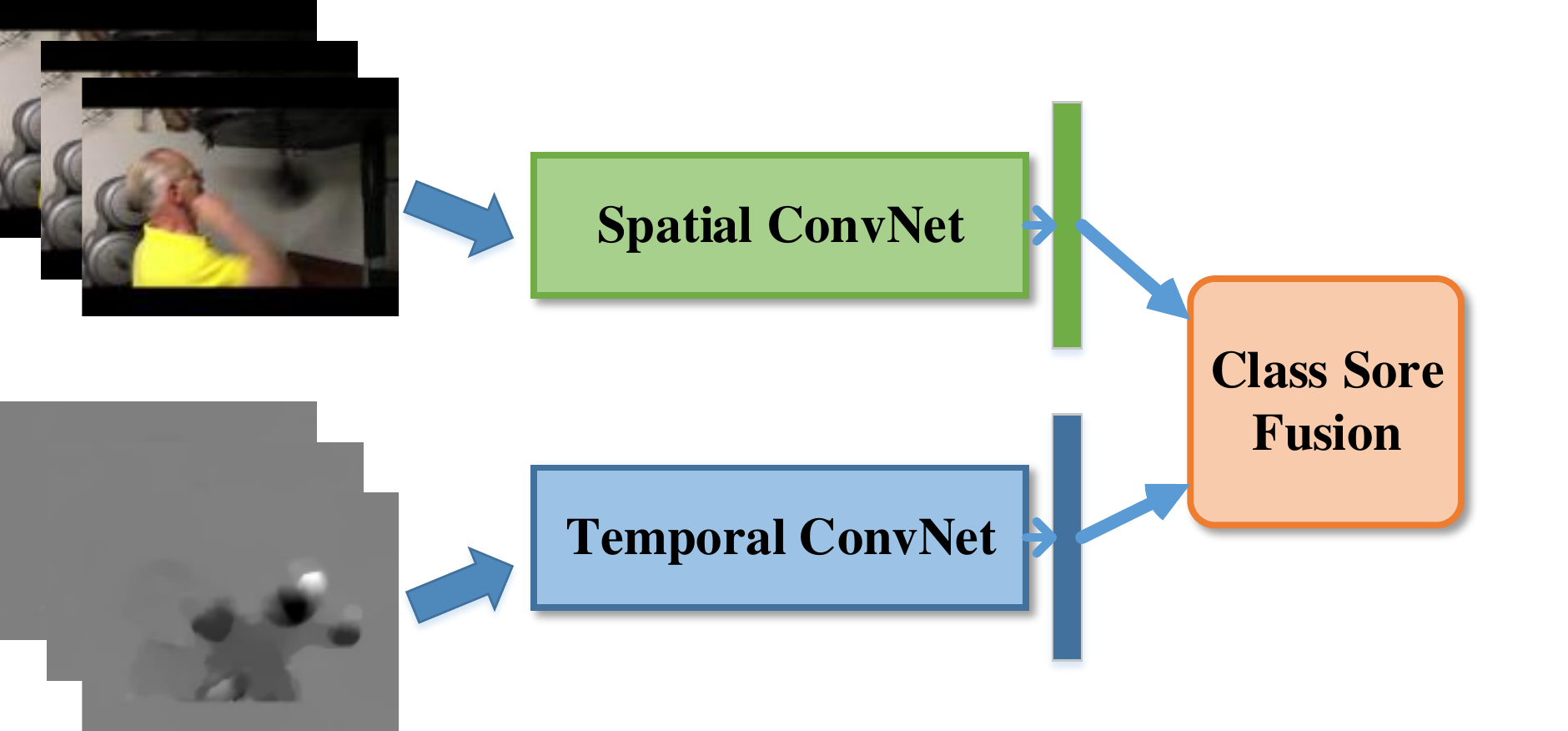}
\caption{The two-stream architecture for action recognition. The spatial ConvNet and temporal ConvNet are trained separately and fused by final classification score.}\label{Fig-network}
\end{figure}

In order to merge these two aspects for action recognition, the two-stream architecture fuses final classification score from two separate ConvNets and achieve state-of-the-art performance in benchmark datasets of human action recognition, as illustrated in Fig. \ref{Fig-network}. Nevertheless, the two-stream architecture cannot exploit these two cues simultaneously, which make it fail when classifying action categories that are ambiguous in individual appearance and motion. For tackling this problem, \cite{feichtenhofer2016convolutional} develops an architecture that is able to fuse spatial and temporal cues at several level feature abstraction. Several intuitive fusion methods, such as sum, max and concatenation, have been studied in this work, but they don't work well due to inconsistent distribution and representation of different modalities (i.e. RGB image and optical flow image). In order to tackle multi-modal information more effectively, we insert a novel message generating and passing component for sharing information between appearance and motion modal, which extracts useful information from one modal and passing it to another one. The passed message have consistent distribution and representation with target modal and provide complementary information to target modal. We summarize the main contributions of this paper as follows:
\begin{itemize}
\item Presenting a novel cross-modal message passing mechanism for two-stream fusion, which alleviates the inconsistent distribution and representation of different modalities and outperforms most intuitive fusion methods.
\item In order to generate meaningful message, an adversarial objective is proposed to train the two-stream network, which promotes competition among different modalities and boost both modal networks simultaneously.
\end{itemize}

\section{Related Work}
\label{sec:rela}
Following the great success of ConvNets for image classification \cite{krizhevsky2012imagenet}, semantic segmentation \cite{long2015fully}, road detection \cite{gao2017conf_embedding} and other computer vision tasks, deep learning algorithms have been used in video-based action recognition as well. Multiple frames in each sequence are feed into ConvNets in \cite{karpathy2014large}, and then pooled using single, late, early and slow fusion across the temporal domain. However, this native method does not obtain significant improvement compared those methods based on a single frame, which indicates that motion features are difficult to obtain by simply and directly pooling spatial features from ConvNets.

In light of this, Karen Simonyan et al. \cite{simonyan2014two} propose a two-stream ConvNet architecture which incorporates spatial and temporal networks, and demonstrate that optical flow is an effective way to model motion information. Since this scheme improves the accuracy of action recognition significantly, the two-stream architecture followed by many other researchers. For example, Bowen Zhang et al. \cite{zhang2016real} accelerate the original two-stream method by replacing optical flow with motion vector which can be obtained directly from compressed videos without extra calculation. Recently, Limin Wang et al. \cite{wang2016temporal} introduce a temporal segment network for long-range temporal structure modeling, which achieves the best performance on benchmark datasets so far. In detail, a sparse temporal sampling strategy is combined with video-level supervision to enable efficient and effective learning using the whole action video. Despite the success of those methods, all of them just simply fuse the final probability scores from two-stream ConvNets, which does not take advantage of complementarity between appearance and motion information when training the network.

In order to fuse the appearance and motion information more effectively, feature-level fusion is a straightforward method. Christoph Feichtenhofer et al. \cite{feichtenhofer2016convolutional} propose a temporal fusion layer that incorporating 3D convolutions and pooling to fuse the feature from the two-stream network, and demonstrate its superiority compared with several simple strategies such as sum, max and concatenation. For leveraging the relation between RGB and corresponding optical flow, Lin Sun et al. \cite{sun2017lattice} extend the traditional LSTM by learning independent hidden state transitions of memory cells for individual spatial locations, and it is trained by both RGB and optical flow and learned weights are shared by both modalities. Moreover, convolutional 3D (C3D) \cite{tran2015learning} is another way to exploit spatial and temporal information jointly, which learns 3D convolution kernels in both space and time based on the straightforward extension of the established 2D CNNs. However, the performance of C3D is inferior to the two-stream architecture method, which verifies the advantage of the two-stream architecture in another way. Therefore, two-stream architecture is the baseline of the proposed method, and a novel message passing mechanism is introduced to harness the appearance and motion information when training and test.

\section{Models and Algorithm}
\label{sec:model}
\subsection{Revisiting Two-stream Architecture}
As illustrated in Fig. \ref{Fig-network}, given a video sequence with $N$ frames, the corresponding optical flow images are extracted as a pre-processing step. For extracting appearance based information, the spatial stream ConvNet operates on a single RGB images, and temporal stream ConvNet takes a stack of $L$ temporally adjacent optical flow images (e.g. $L$=5). The output of spatial and temporal stream ConvNet are $C$-dimensional vector referred as class probability score vector, where $C$ is the number of action category. For fusing those two ConvNets' results, a weighted sum is applied on two output vectors, and final classification result is determined by the max value of the weighted sum vector.In addition, for purpose of modeling the temporal structure of the video, several frames and corresponding optical flow images are sampled by heuristic strategy. For presenting clearly, in the remainder of this paper, we refer the RGB image as Appearance Modal(A-Modal) and the optical flow image as Motion Modal (M-Modal).

\begin{figure}
\centering
\includegraphics[width=.48\textwidth]{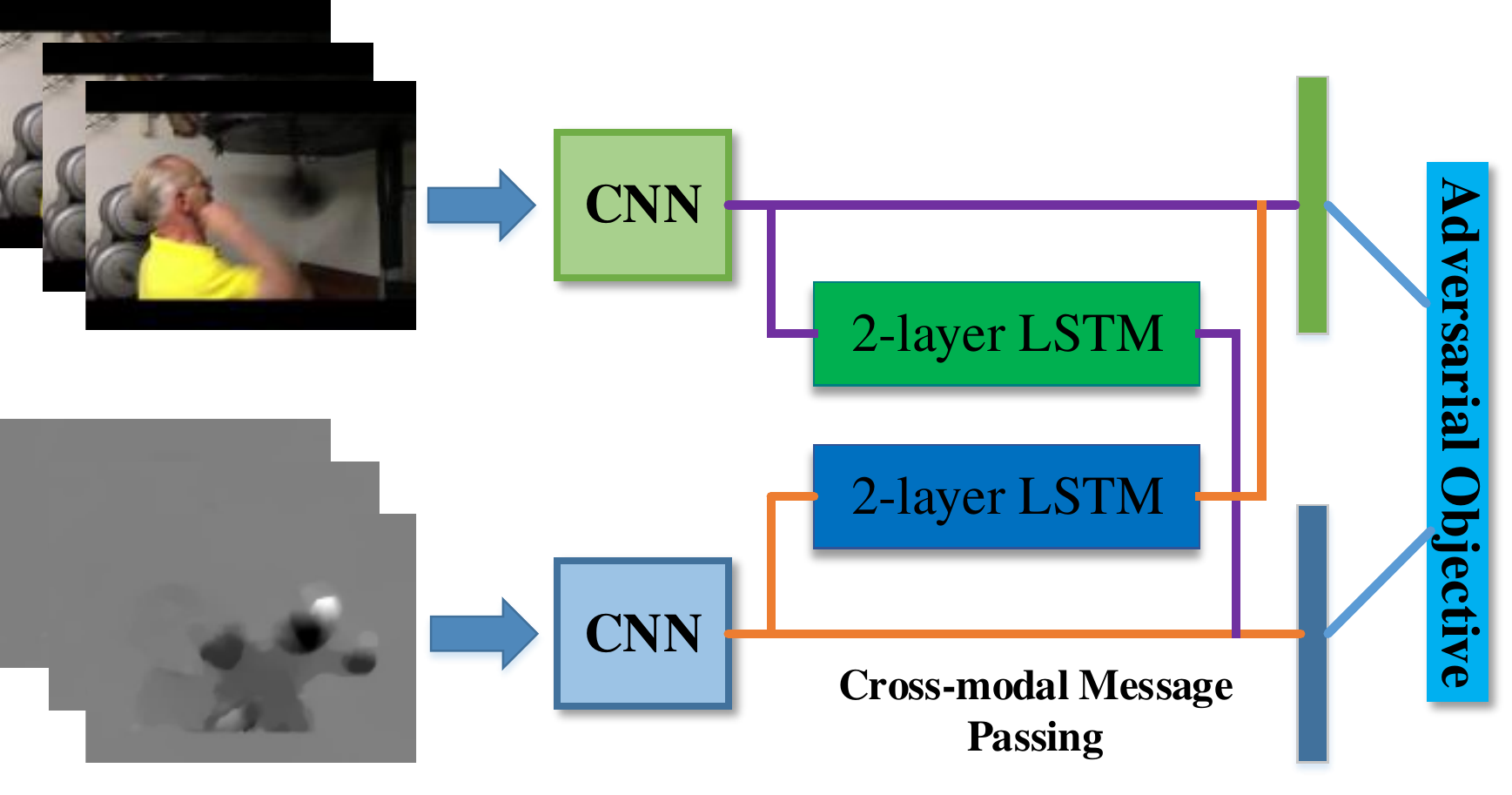}
\caption{Illustration of the proposed CMMP. First, several frames and optical flow images are sampled in a video and feed those into CNNs; Then the extracted feature from two CNNs is regarded as inputs of cross-modal message passing module; Finally the novel adversarial objective is utilized to guide the whole network to update its weight. The whole system is end-to-end trainable.}\label{Fig-our-method}
\end{figure}

\subsection{Cross-modal Message Passing}
As we discussed in Sec. \ref{sec:intro}, an obvious problem of the two-stream ConvNets is their inability in incorporating the appearance and motion information. In detail, the spatial stream ConvNet and temporal stream ConvNet are trained with A-Modal and M-Modal respectively, which make original two-stream architecture can not learn the pixel-wise relationship between spatial and temporal features. For tackling this problem, a novel cross-modal message passing mechanism is proposed to fuse two-stream ConvNets, which alleviates the inconsistent distribution and representation of different modalities and outperforms most intuitive fusion methods. To be specific, two message generator networks are embedded between two-stream ConvNets, where one receives convolutional feature from A-Modal and outputs messages for M-Modal and the other one processes inversely. Both these two message generator are two-layer Long Short Term Memory (LSTM) network but own different weights, as illustrated in Fig. \ref{Fig-our-method}.

Suppose ${{\rm{x}}_a},{{\rm{x}}_m} \in {R^{T \times D}}$ denote the convolutional features from A-Modal and M-Modal respectively, where $T$ is the number of sampled frames and $D$ is the dimensional size of extracted feature. Therefore, the two message generator networks can be formatted as follows:
\begin{equation}
\label{eq-mess-gen}
{\rm{m}_a} = lst{m_2}({\rm{x}_a};{\rm{w}_a});\quad {\rm{m}_m} = lst{m_2}({\rm{x}_m};{\rm{w}_m})
\end{equation}
where $lst{m_2}$ represents the two-layer LSTM and ${\rm{w}_a}, {\rm{w}_m}$ are their weights. Moreover, ${\rm{m}_a},{\rm{m}_m} \in {R^{T \times D}}$ denote the generated messages from A-modal and M-modal, and then those messages are fused with convolutional features from another modal as follows:
\begin{equation}
\label{eq-mess-fuse}
{\rm{x}}_a^f = fusion({x_a},{m_m});\quad {\rm{x}}_m^f = fusion({x_m},{m_a})
\end{equation}
where $fusion$ is a custom function for merging the convolutional features from one modal and message from the other modal, and a simple average strategy is adopted in this paper. For obtaining the classification results, the fused features $\rm{x}_a^f, \rm{x}_m^f$ are then fed into a fully-connected layer, which outputs class probability scores of the input video. In addition, a novel adversarial loss is proposed to ensure that the learned message generator can transform most discriminative information from one modal to the other one.

\subsection{Adversarial Objective}
With the cross-modal message passing mechanism, a fact emerged that a competing objective should to be added in order to make the message generator actually learning meaningful representations of the messages. Hence we introduce an adversarial objective which promotes competition among different modalities and boosts both modal networks simultaneously. Specifically, the adversarial objective is based on the principle that the two-stream ConvNets compete with each other to get lower loss. Following the original two-stream architecture, the standard categorial cross-entropy loss is utilized as loss function for each ConvNets, which is formed as
\begin{equation}
\label{eq-cross-loss}
L(y,s) =  - \sum\limits_{i = 1}^C {{y_i}({s_i} - log\sum\limits_{j = 1}^C {\exp \;{s_j}} )}
\end{equation}
where $C$ is the number of action classes and $y_i$ is the groundtruth label concerning class $i$. Based on this loss function, the adversarial objective function of A-Modal ConvNet is defined as follows:
\begin{equation}
\label{eq-A-ALoss}
A{L_a} = {L_a}(y,{s_a}) + f({L_a}(y,{s_a}) - {L_m}(y,{s_m}))
\end{equation}
while the adversarial objective function for M-Modal ConvNet is:
\begin{equation}
\label{eq-A-ALoss}
A{L_m} = {L_m}(y,{s_m}) + f({L_m}(y,{s_m}) - {L_a}(y,{s_a}))
\end{equation}
where $L_a, L_m$ represent the cross-entropy loss of A-Modal and M-Modal ConvNets; $f(x)= max(x,0)$ so that the optimization objective for A-Modal ConvNet pushes it to get lower loss and vice versa for M-Modal ConvNet.

\subsection{Training Details}
\textbf{Two-Stage Training:} The training frameworks for the proposed cross-modal message passing network comprises of two stages. First, two-stream ConvNets is pretrained using standard categorial cross-entropy loss without updating the cross-modal message passing network, and then the proposed adversarial objective loss function is utilized to train the whole network jointly. In pretraining stage, the ConvNet is trained to predict action category only using one modal data. In this way, each ConvNet learns to extract the most relevant feature from each modal for action classification. In other words, the spatial-stream ConvNet will capture the appearance characteristics about action, such as skis in skiing, soccer ball in playing football. Meanwhile, the spatial-stream ConvNet will exploit the motion characteristics about action, for example, difference between walking and running, yawning and laughing, or swimming, crawl and breast-stroke.

Due to inconsistent distribution between A-Modal and M-Modal, the learned feature representation also has different distributions and simple fusion techniques (such as sum, max and concatenation) is not effective. Therefore, after pretraining, we fine-tune the proposed cross-modal message passing network with a novel adversarial objective. Compared with the standard categorial cross-entropy loss, the proposed adversarial objective function forces the message generator network transform the most discriminative feature from one source modal to another target one, which boosts the target modal ConvNet achieves better performance. The effectiveness of this adversarial objective is demonstrated in Table \ref{Table-HDMB-1}.

\section{Experiments}
\label{sec:experiment}
\subsection{Experiments Setup}
Experiments are mainly conducted on two action recognition benchmark datasets, namely UCF-101 \cite{soomro2012ucf101} and HMDB-51 \cite{kuehne2011hmdb}, which are currently the largest and most challenging annotated action recognition datasets. The UCF-101 dataset consists of 13320 action videos in 101 categories and the HMDB-51 contains 6766 videos that have been annotated for 51 actions. For both datasets, we use the same training strategy and report average accuracy according to the provided evaluation protocol.

The mini-batch stochastic gradient descent algorithm is utilized to learn the network parameters, where the batch size is set to 64 and momentum set to 0.9. We initialize the network weights with pre-trained models from ImageNet. The pretraining learning rate is initialized as 0.001 and the fine-tuning learning rate is initialized as 0.0001, and both of learning rate decreases to its $1/10$ every 4500 iterations. The whole training procedure stops at 13500 iterations.

\subsection{Exploration Study}
In this section, we list the performance of several simple fusion methods and our main components for action recognition on split 1 of HMDB-51 in Table \ref{Table-HDMB-1}. Specifically, we compare different fusion method without adversarial objective, referred as SUM, MAX and CMMP+noAL. we see that the final fusion result of CMMP+noAL is better than SUM and MAX , which demonstrates that proposed cross-modal message passing mechanism is more effective for two-stream fusion.

In addition, we also analyze the effectiveness of the proposed adversarial objective and two-stage training strategy. Specifically, we compare two kind of settings: without and with adversarial objective. The results are summarized in Table \ref{Table-HDMB-1}. First, we see that the performance of MAX+AL  is a little better than AL for spatial ConvNets, temporal ConvNets and Fusion result. Then, we resort to the two-stage training framework to help initialize CMMP and achieve better performance using the proposed adversarial objective. As shown in Table \ref{Table-HDMB-1}, the proposed adversarial objective boosts the recognition performance by a significant margin, especially for temporal ConvNets, which implies that the message from A-Modal to M-Modal is very necessary.
\begin{table}[htbp]
\centering
\caption{CMMP components analysis on split 1 of HMDB-51.}
\begin{tabular}{c|c|c|c}
\hline
Method &Spatial &Temporal &Fusion \\ \hline
SUM &53.01 &54.05 &53.79 \\
MAX &52.61 &52.29 &52.68 \\
CMMP+noAL &46.99 &47.71 &60.13 \\  \hline
\hline
SUM+AL &51.70 &52.29 &51.96 \\
MAX+AL &53.79 &53.66 &53.88 \\
CMMP &\textbf{50.07} &\textbf{65.23} &\textbf{66.67} \\ \hline
\end{tabular}\label{Table-HDMB-1}
\end{table}

\subsection{Comparison with the-state-of-the-art}
After exploring the effect of components in the proposed method, we compare our method with many other approaches and the results are summarized in Table \ref{Table-ALL}. In detail, we compare the proposed method with traditional methods such as improved trajectories (iDTs) \cite{wang2013action} and deep learning approaches such as two-stream networks \cite{NIPS2014_5353}, factorized spatio-temporal convolutional networks \cite{sun2015human}, 3D convolutional networks (C3D) \cite{tran2015learning}, trajectory-pooled deep convolutional descriptors \cite{wang2015action}, and spatio-temporal fusion CNNs \cite{feichtenhofer2016convolutional}.
\begin{table}[htbp]
\centering
\caption{Mean accuracy on the UCF-101 and HMDB-51.}
\begin{tabular}{|c|c|c|c|}
\hline
Model &Method &UCF-101 &HMDB-51 \\ \hline
\multirow{2}{*}{Traditional} &iDT+FV \cite{wang2013action} &85.9 &57.2 \\
&iDT+HSV \cite{peng2016bag} &87.9 &61.6 \\ \hline
\hline
\multirow{7}{*}{Deep} &EMV-CNN \cite{zhang2016real} &86.4 &- \\
&Two Stream \cite{NIPS2014_5353} &88.0 &59.4 \\
&$F_{ST}$CN \cite{sun2015human} &88.1 &59.1 \\
&C3D \cite{tran2015learning} &85.2 &- \\
&VideoLSTM \cite{li2016videolstm} &89.2 &56.4 \\
&TDD+FV \cite{wang2015action} &90.3 &63.2 \\
&Fusion \cite{feichtenhofer2016convolutional} &91.8 &64.6 \\ \hline
Ours &CMMP &\textbf{91.3} &\textbf{65.9} \\ \hline
\end{tabular}\label{Table-ALL}
\end{table}

\section{Conclusion}
\label{sec:conclusion}
In order to fuse two-stream ConvNets more effectively for action recognition, CMMP is proposed. This method transfers the discriminative message from one modal to another, which alleviates the inconsistent distribution and representation of different modalities. We also introduced a novel adversarial objective to fine-tune the whole network, and boosts the performance even further. A huge gain is observed compared with other simple fusion methods, and a comparable performance is achieved in whole benchmark datasets.

\bibliographystyle{IEEEbib}
\bibliography{cmmp}

\end{document}